\documentclass[conference]{IEEEtran}
\IEEEoverridecommandlockouts
\usepackage{dblfloatfix}
\usepackage{cite}
\usepackage{amsmath,amssymb,amsfonts}
\usepackage{textcomp}
\usepackage[table,xcdraw,dvipsnames]{xcolor}
\usepackage[hidelinks,colorlinks=true,linkcolor=blue,citecolor=blue,breaklinks=true]{hyperref}
\usepackage[T1]{fontenc}
\usepackage[utf8]{inputenc}
\usepackage{graphicx}
\usepackage{caption}
\usepackage{multirow}
\usepackage{makecell}
\usepackage{colortbl}
\usepackage{booktabs}
\usepackage{array}
\usepackage{tabularx}
\usepackage{url}
\usepackage{tikz}
\usetikzlibrary{arrows.meta,positioning}
\graphicspath{{./}{figures/}}

\definecolor{lightpink1}{HTML}{FCF4F6}
\definecolor{lightpink2}{HTML}{FBECEF}
\definecolor{lightblue1}{HTML}{D6ECFF}
\definecolor{lightblue2}{HTML}{E9F4FF}
\definecolor{headbg}{HTML}{FFF4F5}
\definecolor{medbg}{HTML}{F2FBF4}
\definecolor{tailbg}{HTML}{F3F7FF}
\definecolor{metricdelta}{HTML}{9A3A3A}
\definecolor{pubyearcolor}{HTML}{777777}
\newcommand{\metricchange}[1]{{\fontsize{4.8}{5.2}\selectfont\bfseries\textcolor{metricdelta}{(#1)}}}
\newcommand{\pubyear}[1]{\textcolor{pubyearcolor}{(#1)}}

\title{SHTA: Semantic Hard Token Correction and Center Alignment for Semi-Supervised Medical Image Segmentation}

\author{

\IEEEauthorblockN{
Zhuoru Zhang\textsuperscript{2},
Yiheng Zhong\textsuperscript{1},
Zimu Zhang\textsuperscript{2},
Xiaofeng Liu\textsuperscript{1,\dag}
}

\IEEEauthorblockA{
\textsuperscript{1}Yale University, United States\\
\textsuperscript{2}Xi'an Jiaotong-Liverpool University, China\\
Emails: xiaofeng.liu@yale.edu
}

\thanks{\textsuperscript{\dag}Corresponding authors.}
}

\begin{document}
\maketitle
\begin{abstract}
Recent advances in semi-supervised medical image segmentation have achieved remarkable performance through prediction consistency, pseudo-label supervision, and hard-region supervision. However, these methods primarily improve supervision quality rather than explicitly enforcing semantic consistency in the learned representations of hard regions. Consequently, even under increasingly stronger prediction-level supervision, difficult regions exhibiting unstable semantic assignment often fail to establish semantically consistent representations during training, thereby limiting further segmentation improvement. To address this issue, we propose \textbf{SHTA} (\textbf{S}emantic \textbf{H}ard \textbf{T}oken Correction and Center \textbf{A}lignment), a lightweight training-time semantic representation branch. Instead of introducing additional prediction supervision, SHTA refines intermediate semantic representations through Semantic Assignment, Hard Token Refinement, and Semantic Center Alignment, thereby improving semantic consistency in hard regions while preserving the original prediction pathway and introducing no additional inference cost. We integrate SHTA into representative semi-supervised segmentation frameworks, including GA-CPS, CPS, URPC, and MagicNet, and conduct evaluations on the Synapse and AMOS datasets. Experimental results demonstrate that SHTA delivers consistent paired improvements across frameworks, with especially clear gains in segmentation accuracy, weak-organ recovery, and semantic ambiguity reduction, while incurring only training-time overhead. The code is available at \url{https://anonymous.4open.science/r/release_SHTA-42D5/}.
\end{abstract}


\section{Introduction}

Anatomical organ segmentation is a fundamental task in medical image analysis and plays an important role in diagnosis assistance, treatment planning, and quantitative clinical assessment. In 3D abdominal CT segmentation, models are required to delineate multiple organs from volumetric scans, where accurate voxel-wise predictions are essential for reliable clinical interpretation. Although deep learning has achieved remarkable progress in medical image segmentation~\cite{ref_dlseg_1,ref_dlseg_2,zhong2025pg,ref_dlseg_3}, dense voxel-level annotation remains expensive and time-consuming, especially when multiple anatomical structures must be labeled by trained experts.

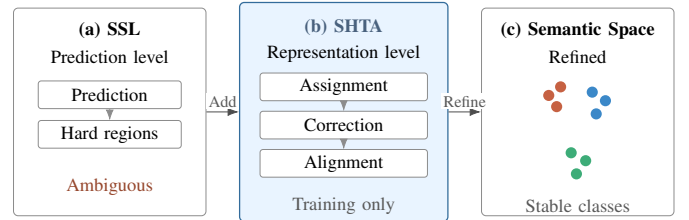
\begin{figure}[!tbp]
  \centering
  \resizebox{\columnwidth}{!}{
  \begin{tikzpicture}[
      font=\footnotesize,
      panel/.style={draw=black!45, rounded corners=2pt, minimum width=2.92cm, minimum
      height=3.15cm, align=center, fill=white},
      mainpanel/.style={draw=RoyalBlue!70!black, line width=0.55pt, rounded corners=2pt,
      minimum width=3.16cm, minimum height=3.32cm, align=center, fill=lightblue2},
      title/.style={font=\bfseries\footnotesize, align=center, inner sep=1pt},
      note/.style={font=\footnotesize, align=center, inner sep=1pt},
      box/.style={draw=black!45, rounded corners=1.3pt, minimum width=2.12cm, minimum
      height=0.42cm, fill=white, align=center, text width=2.05cm, inner sep=1.7pt},
      arrowlabel/.style={font=\scriptsize, fill=white, inner sep=1pt, text=black!65},
      arr/.style={-Latex, line width=0.42pt, draw=black!70},
      weakarr/.style={-Latex, line width=0.32pt, draw=black!45}
  ]

  \node[panel] (ssl) at (0,0) {};
  \node[mainpanel] (shta) at (3.55,0) {};
  \node[panel] (out) at (7.10,0) {};

  \node[title] at (0,1.24) {\textbf{(a)} SSL};
  \node[note] at (0,0.84) {Prediction level};
  \node[box] (pred) at (0,0.26) {Prediction};
  \node[box] (pseudo) at (0,-0.34) {Hard regions};
  \draw[weakarr] (pred) -- (pseudo);

  \node[note, text=BrickRed!75!black] at (0,-1.14) {Ambiguous};

  \node[title, text=RoyalBlue!45!black] at (3.55,1.28) {\textbf{(b)} SHTA};
  \node[note] at (3.55,0.88) {Representation level};

  \node[box, minimum width=2.50cm, text width=2.42cm] (org) at (3.55,0.38) {Assignment};
  \node[box, minimum width=2.50cm, text width=2.42cm] (corr) at (3.55,-0.20) {Correction};
  \node[box, minimum width=2.50cm, text width=2.42cm] (stab) at (3.55,-0.78) {Alignment};
  \draw[weakarr] (org) -- (corr);
  \draw[weakarr] (corr) -- (stab);

  \node[note, text=black!65] at (3.55,-1.42) {Training only};

  \node[title] at (7.10,1.24) {\textbf{(c)} Semantic Space};
  \node[note] at (7.10,0.84) {Refined};

  \fill[BrickRed!75] (6.66,0.25) circle (2.4pt);
  \fill[BrickRed!75] (6.84,0.38) circle (2.4pt);
  \fill[BrickRed!75] (6.78,0.08) circle (2.4pt);
  \fill[RoyalBlue!75] (7.32,0.30) circle (2.4pt);
  \fill[RoyalBlue!75] (7.52,0.17) circle (2.4pt);
  \fill[RoyalBlue!75] (7.38,-0.03) circle (2.4pt);
  \fill[ForestGreen!75] (7.00,-0.62) circle (2.4pt);
  \fill[ForestGreen!75] (7.22,-0.74) circle (2.4pt);
  \fill[ForestGreen!75] (7.08,-0.94) circle (2.4pt);

  \node[note, text=black!65] at (7.10,-1.42) {Stable classes};

  \draw[arr] (ssl.east) -- node[arrowlabel, above=1pt] {Add} (shta.west);
  \draw[arr] (shta.east) -- node[arrowlabel, above=1pt] {Refine} (out.west);

  \end{tikzpicture}}
  \caption{Conceptual comparison between prediction-level SSL and SHTA. \textbf{(a)} Conventional SSL selects prediction-level hard evidence. \textbf{(b)} SHTA performs representation-level assignment, correction, and alignment. \textbf{(c)} Corrected hard-region tokens form a more stable semantic space.}
  \label{fig:conceptual_control}
\end{figure}

Semi-supervised medical image segmentation alleviates this annotation burden by learning from a small set of labeled volumes together with abundant unlabeled data. Existing SSL methods mainly improve how unlabeled data are used for supervision. Prediction-consistency methods enforce agreement across teacher--student networks, dual branches, or perturbed views~\cite{ref_ssl_1,ref_ssl_2,ref_ssl_3,ref_ssl_4}, while reliable-supervision methods refine pseudo labels, mine hard regions, estimate uncertainty, or select reliable samples~\cite{bai2023bcp,chen2023magicnet,chi2024abd,wang2023dhc,yin2025skcdf,ref_imbalance_4,wang2025dca}. These strategies are effective for deciding which predictions, pseudo labels, samples, or regions should supervise training. However, as summarized in Fig.~\ref{fig:conceptual_control}(a), they still mainly operate at the prediction or region-selection level; after a hard region is selected, the semantic organization of its intermediate token representations is only indirectly constrained.

This post-selection ambiguity is most pronounced for small, thin, or boundary-adjacent organs. Such regions often contain limited voxels, have weak intensity contrast, and are easily confused with adjacent anatomical structures. Therefore, even when a hard region has been selected as useful training evidence, its token embedding may still lie close to competing classes in the representation space. This causes unstable token-to-class assignment and weak class-consistent aggregation, meaning that the selected evidence is useful at the supervision level but still ambiguous at the semantic-representation level.

This observation motivates our central question: \textbf{after difficult evidence has been selected, how can its token-to-class assignment be corrected and its class-level semantic structure be stabilized?} We argue that semi-supervised hard-region learning needs post-selection semantic correction, not only better feature discriminability or stronger prediction-level supervision. Reliable labeled masks provide a natural semantic reference for this purpose, but they must be converted into token-level guidance that can act on selected hard-region representations.

Inspired by representation-level methods that organize latent features through prototypes, proxies, or auxiliary semantic branches~\cite{cpcl2021,scpnet2023,mpcl2024,pccs2025,ref_semantic_module_1}, we propose SHTA, a lightweight training-only semantic representation branch for semi-supervised 3D medical image segmentation. SHTA converts labeled masks into token-level semantic guidance, corrects hard-token assignments after hard-region selection, and stabilizes class-level representation geometry. Specifically, as shown in Fig.~\ref{fig:conceptual_control}(b), SHTA first organizes token embeddings through proxy-based Semantic Assignment, then uses labeled-token semantic guidance in Hard Token Refinement to refine ambiguous hard-token assignments, and finally aggregates the corrected hard tokens into class-level centers for Semantic Center Alignment. This produces the refined semantic space illustrated in Fig.~\ref{fig:conceptual_control}(c), where selected hard-region tokens become better aligned with their class semantics. The detailed architecture is shown in Fig.~\ref{fig:framework}. The branch is attached only during training and removed entirely during inference, without changing the base segmentation architecture or adding inference cost.

The main contributions of this work are summarized as follows:

\begin{itemize}
    \item We formulate the post-selection semantic ambiguity problem in semi-supervised medical image segmentation, where selected hard evidence may still have unstable token-to-class assignments and weak class-consistent representations.

    \item We propose SHTA, a plug-and-play training-only semantic branch that converts labeled masks into token-level semantic guidance, corrects selected hard-token assignments, and aligns class-level semantic centers through Semantic Assignment, Hard Token Refinement, and Semantic Center Alignment.

    \item We validate SHTA on the Synapse and AMOS benchmarks across representative SSL frameworks, demonstrating consistent paired-baseline improvement together with ablation, semantic-consistency, and computational-overhead analyses.
\end{itemize}

\section{Related Work}

\subsection{Prediction-Level Consistency}

Prediction-level consistency is one of the most widely adopted paradigms for semi-supervised medical image segmentation. Representative methods such as UA-MT, CPS, and SS-Net enforce prediction agreement across teacher--student networks, dual branches, or perturbed views, while URPC further introduces uncertainty-aware consistency regularization to improve robustness under limited annotations~\cite{ref_ssl_1,ref_ssl_2,ref_ssl_3,ref_ssl_4}. In general, these methods use unlabeled data by encouraging stable output masks or logits under model, data, or perturbation differences, thereby regularizing the final prediction space from limited labeled data.

\subsection{Reliable Supervision}

Another important direction improves the quality of supervision under limited annotations. Representative methods, including BCP, MagicNet, ABD, DHC, SKCDF, GA, and DCA, enhance learning through pseudo-label refinement, class-aware weighting, hard-region mining, and reliable sample selection strategies~\cite{bai2023bcp,chen2023magicnet,chi2024abd,wang2023dhc,yin2025skcdf,ref_imbalance_4,wang2025dca}. These techniques are commonly used in multi-organ segmentation, where class imbalance, small organs, and ambiguous boundaries make supervision noisy. From a supervision perspective, they help determine which pseudo labels, regions, classes, or samples should contribute more reliable training signals.

\subsection{Semantic Representation Learning}

Recent studies have explored representation-level learning to improve feature discriminability by explicitly organizing semantic structures in the latent feature space. Representative methods, including CPCL, SCP-Net, MPCL, and PCCS, introduce prototype- or proxy-based supervision to encourage compact intra-class representations and improved inter-class separability~\cite{cpcl2021,scpnet2023,mpcl2024,pccs2025}. In addition, ICL incorporates auxiliary semantic representation branches during training, which can be removed during inference while remaining compatible with existing SSL frameworks~\cite{ref_semantic_module_1}. These studies provide useful tools for structuring intermediate representations. In this context, our work studies a specific post-selection setting: after difficult evidence has been identified, its semantic assignment and class-level structure are refined using labeled-token guidance.

\section{Method}
\subsection{Overview}
\begin{figure*}[!tbp]
\centering
\includegraphics[width=\textwidth]{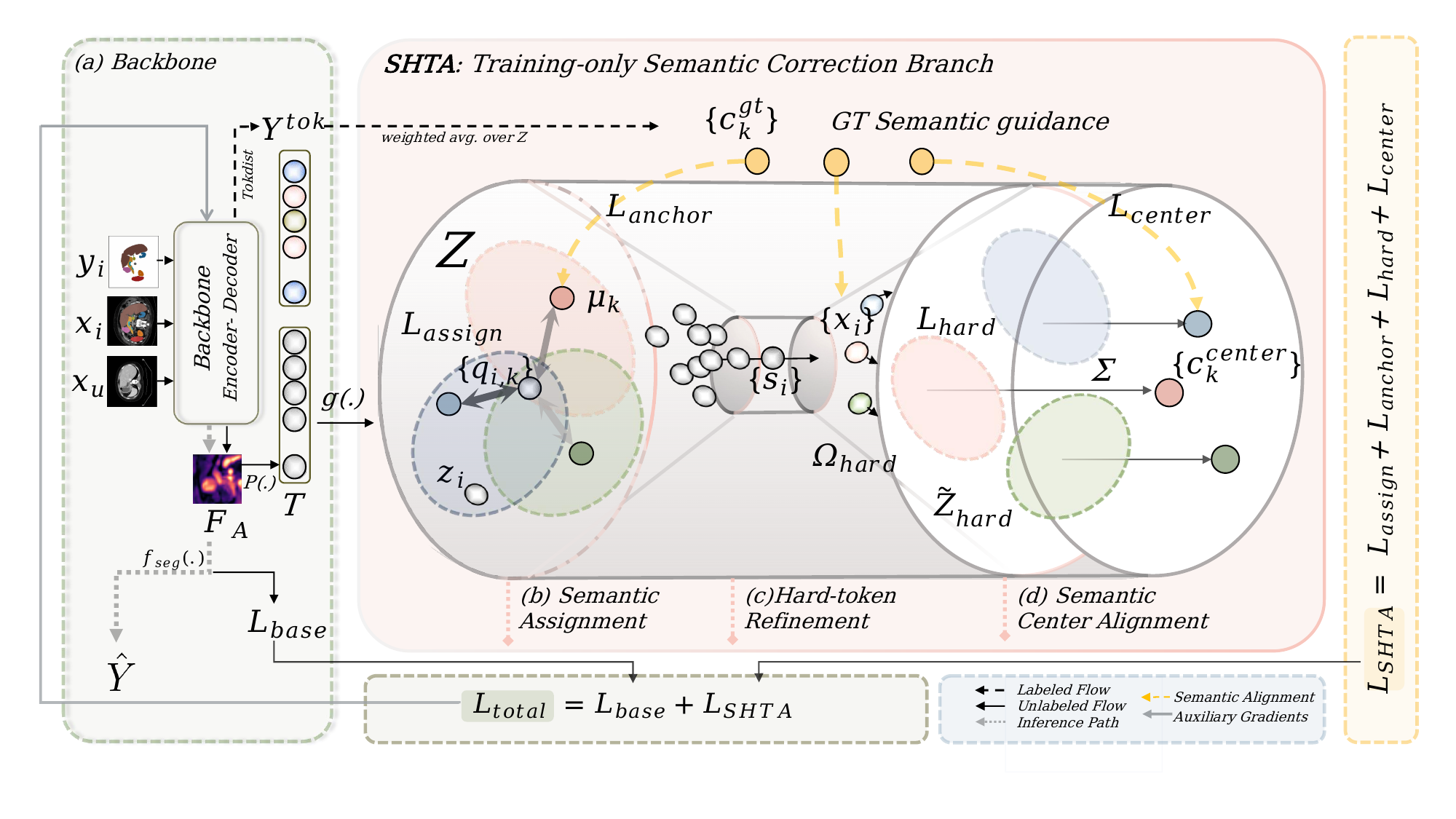}
\caption{Overview of the SHTA semantic branch. SHTA contains (b) Semantic Assignment, (c) Hard Token Refinement, and (d) Semantic Center Alignment. Solid, dotted, and yellow arrows denote training flow, inference path, and semantic-alignment paths, respectively.}
\label{fig:framework}
\end{figure*}
  SHTA is attached to an existing SSL segmenter as an auxiliary semantic branch. It taps the
  decoder feature \(F_A\) and converts labeled masks into token-level guidance for three operations:
  token organization, hard-token correction, and class-center stabilization.

  Formally, let \(\mathcal{D}_L=\{(x_l,y_l)\}\) and \(\mathcal{D}_U=\{x_u\}\) be the labeled and
  unlabeled training sets, and let \(Y_L=\{y_l\}\) be the labeled masks. The base framework
  produces a decoder feature \(F_A \in \mathbb{R}^{B\times C\times D\times H\times W}\).
  The original prediction path maps this feature to the segmentation prediction
  \(\hat{Y}=f_{\mathrm{seg}}(F_A)\) and is optimized by the base objective
  \(\mathcal{L}_{base}\). Here, \(B\), \(C\), and \(D\times H\times W\) denote the batch size,
  channel dimension, and spatial resolution, respectively. SHTA does not feed into
  \(f_{\mathrm{seg}}\), alter \(\mathcal{L}_{base}\), or modify the labeled/unlabeled training
  flows of the base SSL framework.

  The semantic branch first constructs the token sequence, semantic embeddings, and token-level
  labeled semantic distributions as supporting objects:
  \begin{equation}
  \begin{aligned}
  T &= P(F_A), \qquad Z = g(T),\\
  Y^{\mathrm{tok}} &= \mathrm{TokDist}(Y_L).
  \end{aligned}
  \label{eq:token_objects}
  \end{equation}
  Here, \(P(\cdot)\) denotes token projection, \(g(\cdot)\) is the
  semantic embedding head, \(T=\{t_i\}_{i=1}^{L}\in\mathbb{R}^{B\times L\times C}\) is the full
  token sequence, and \(Z=\{z_i\}_{i=1}^{L}\in\mathbb{R}^{B\times L\times C_z}\) is the
  semantic embedding sequence, where \(C_z\) is the semantic embedding dimension.
  \(\mathrm{TokDist}(\cdot)\) converts labeled masks into token-level class proportions, yielding
  \(Y^{\mathrm{tok}}\in\mathbb{R}^{B\times L\times K}\). Here,
  \(L\) is the number of tokens, and \(K\) is the number of semantic classes including background.
  We use \(k\in\{0,\ldots,K-1\}\) as the class index, with \(k=0\) denoting background and
  \(k>0\) denoting foreground classes. \(Y^{\mathrm{tok}}_{i,k}\) denotes the proportion of class
  \(k\) in token \(i\), and \(\Omega_l\) denotes the labeled-token index set.

  Given \((Z,Y^{\mathrm{tok}})\), SHTA has three modules: Semantic Assignment
  (Fig.~\ref{fig:framework}(b)) organizes tokens with learnable class proxies, Hard Token
  Refinement (Fig.~\ref{fig:framework}(c)) corrects selected hard-token assignments, and
  Semantic Center Alignment (Fig.~\ref{fig:framework}(d)) stabilizes class-level semantic
  centers. Their losses are combined with the base objective as:
  \begin{equation}
  \mathcal{L}_{total}
  =
  \mathcal{L}_{base}
  + \lambda_{shta}\mathcal{L}_{\mathrm{SHTA}} .
  \label{eq:total_objective}
  \end{equation}
  Here, \(\lambda_{shta}\) controls the contribution of the auxiliary objective, and
  \(\mathcal{L}_{\mathrm{SHTA}}\) is defined after the three modules.

\subsection{Semantic Assignment via Proxy Calibration}

Semantic Assignment builds class-aware token organization before hard-token correction.
We introduce a learnable proxy \(\mu_k\in\mathbb{R}^{C_z}\) for each class \(k\), and denote the normalized proxy bank by
\(\bar{\mathbf{M}}=[\bar{\mu}_0,\ldots,\bar{\mu}_{K-1}]^\top\), where
\(\bar{\mu}_k=\mu_k/\|\mu_k\|_2\).

For each token embedding \(z_i\), we compute its cosine similarity to each class proxy and
normalize the scores into a token-to-class assignment:
\begin{equation}
\begin{aligned}
s_{i,k} &= \frac{\bar{\mu}_k^\top \bar{z}_i}{\tau_{\mathrm{sa}}},\\
\mathbf{q}_i &= \operatorname{softmax}(\mathbf{s}_i),
\qquad q_{i,k}=(\mathbf{q}_i)_k .
\end{aligned}
\label{eq:proxy_assignment}
\end{equation}
where \(\bar{z}_i=z_i/\|z_i\|_2\), \(\mathbf{s}_i=[s_{i,0},\ldots,s_{i,K-1}]\),
and \(\tau_{\mathrm{sa}}\) is the semantic-assignment temperature. The scalar \(q_{i,k}\) is the
central assignment variable used in subsequent semantic-assignment supervision and hard-token selection,
indicating how strongly token \(i\) is associated with class \(k\).

Since learnable proxies may drift during training, we also compute the labeled-token class
reference and its valid class set:
\begin{equation}
\begin{aligned}
c_k^{gt}
&=
\frac{\sum_{i\in\Omega_l}Y^{\mathrm{tok}}_{i,k}z_i}
{\sum_{i\in\Omega_l}Y^{\mathrm{tok}}_{i,k}+\epsilon},\\
\mathcal{K}'
&=
\left\{k\in\{0,\ldots,K-1\} \mid
\sum_{i\in\Omega_l}Y^{\mathrm{tok}}_{i,k}>0
\right\}.
\end{aligned}
\label{eq:gt_reference}
\end{equation}
Here, \(c_k^{gt}\) is the GT-derived class reference shown in Fig.~\ref{fig:framework}, computed
only from labeled tokens that contain class \(k\). \(\mathcal{K}'\) is the set of classes present
in the labeled tokens, and \(\epsilon\) is a small constant for numerical stability.

The Semantic Assignment objective is:
\begin{equation}
\begin{aligned}
\mathcal{L}_{assign}
&=
-\frac{1}{|\Omega_l|}
\sum_{i\in\Omega_l}\sum_{k=0}^{K-1}
Y^{\mathrm{tok}}_{i,k}\log(q_{i,k}+\epsilon),\\
\mathcal{L}_{anchor}
&=
\frac{1}{|\mathcal{K}'|}
\sum_{k\in\mathcal{K}'}
\left(1-\cos(c_k^{gt},\mu_k)\right).
\end{aligned}
\label{eq:assign_anchor_loss}
\end{equation}
\(\mathcal{L}_{assign}\) provides token-to-class assignment supervision, while
\(\mathcal{L}_{anchor}\) is the proxy calibration regularizer.
Here, \(\cos(\cdot,\cdot)\) denotes cosine similarity. Together, these two losses provide the
semantic scaffold for Hard Token Refinement.

\subsection{Hard Token Refinement}

Hard Token Refinement corrects selected hard tokens using the GT token distribution of labeled
samples. Starting from Eq.~\eqref{eq:proxy_assignment}, it selects reliable foreground hard tokens
and supervises them toward their dominant GT foreground classes.

Hard-token selection is supported by three token-wise quantities: semantic confidence \(h_i\),
foreground ratio \(m_i^{fg}\), and foreground-class purity \(p_i^{fg}\). For token \(i\), they are
defined as:
\begin{equation}
\begin{aligned}
h_i &= \max_{0\leq k<K} q_{i,k},\\
m_i^{fg} &= \sum_{k>0}Y^{\mathrm{tok}}_{i,k},\\
p_i^{fg} &= \max_{k>0}Y^{\mathrm{tok}}_{i,k}.
\end{aligned}
\label{eq:hard_scores}
\end{equation}
where \(q_{i,k}\) is the token-to-class assignment from Semantic Assignment,
\(Y^{\mathrm{tok}}_{i,k}\) is the token-level GT class proportion, and \(k=0\) denotes background;
therefore, \(k>0\) indexes foreground classes.

Using these scores, SHTA builds a candidate hard-token set \(\Omega_{cand}\) and then selects the
final hard-token set \(\Omega_{hard}\):
\begin{equation}
\begin{aligned}
\Omega_{cand} =
\left\{
i\in\Omega_l \mid h_i \ge T_{conf},\;
m_i^{fg} \ge T_{fg},\;
p_i^{fg} \ge T_{pur}
\right\},\\
\Omega_{hard}=\operatorname{Top}_r(\Omega_{cand}; h_i).
\end{aligned}
\label{eq:hard_sets}
\end{equation}
Here, \(\Omega_l\) is the labeled-token index set, and \(T_{conf}\), \(T_{fg}\), and \(T_{pur}\)
are the confidence, foreground-ratio, and purity thresholds, respectively.
\(\operatorname{Top}_r(\Omega_{cand};h_i)\) selects the top-\(r\) candidates ranked by \(h_i\),
where \(r\) denotes the hard-token selection ratio. In Fig.~\ref{fig:framework}(c),
\(\{s_j\}\) denotes an enumeration of the selected hard-token indices in \(\Omega_{hard}\);
the displayed hard-token embeddings \(\{x_j\}\) are obtained as \(x_j=z_{s_j}\).

For correction, each selected hard token uses the dominant foreground class in its GT token
distribution as the target, i.e., \(y_i^{hard}=\arg\max_{k>0}Y^{\mathrm{tok}}_{i,k}\). The
selected embedding is denoted as \(x_i=z_i\); through the correction loss below, its assignment
and embedding are updated by gradients toward this hard-token target.

The Hard Token Refinement loss is:
\begin{equation}
\mathcal{L}_{hard}=
\frac{1}{|\Omega_{hard}|}
\sum_{i\in\Omega_{hard}}
\mathrm{CE}(\mathbf{q}_i,y_i^{hard}) .
\label{eq:hard_loss}
\end{equation}
This loss corrects the proxy assignment of selected hard foreground tokens, where
\(\mathrm{CE}(\cdot)\) denotes cross-entropy and \(\mathbf{q}_i\) is the assignment vector of token
\(i\). After this gradient-based correction, the selected embeddings are denoted as
\(\tilde{Z}_{hard}=\{\tilde{z}_i\mid i\in\Omega_{hard}\}\) and are passed to Semantic Center
Alignment for class-level semantic stabilization.

\subsection{Semantic Center Alignment}

Semantic Center Alignment aggregates refined hard-token embeddings into class-wise semantic centers.
For each foreground class \(k\), let
\(\Omega_k=\{i\in\Omega_{hard}\mid y_i^{hard}=k\}\) denote the selected hard tokens assigned to
class \(k\). We define the corrected class-level semantic center \(c_k^{center}\) using the
aggregation operator \(\Sigma\) in Fig.~\ref{fig:framework}, implemented as the following masked
average:
\begin{equation}
c_k^{center}
=
\frac{1}{|\Omega_k|}
\sum_{i\in\Omega_k}\tilde{z}_i .
\label{eq:center_definition}
\end{equation}
Here, \(\tilde{z}_i\) is the refined embedding of selected hard token \(i\). We use
\(\mathcal{K}_h=\{k>0\mid \Omega_k\neq\emptyset\}\) to denote foreground classes containing
selected hard tokens.

The center-alignment loss aligns these centers with the GT-derived references from
Eq.~\eqref{eq:gt_reference}:
\begin{equation}
\mathcal{L}_{center}
=
\frac{1}{|\mathcal{K}_h|}
\sum_{k\in\mathcal{K}_h}
\left(1-\cos(c_k^{center},c_k^{gt})\right).
\label{eq:center_loss}
\end{equation}
This loss aligns locally corrected token representations with class-level semantic references.

\subsection{Total Loss}

The final SHTA objective combines the three module losses as:
\begin{equation}
\mathcal{L}_{\mathrm{SHTA}} =
\alpha_{as}\mathcal{L}_{assign}
+ \alpha_{a}\mathcal{L}_{anchor}
+ \alpha_{h}\mathcal{L}_{hard}
+ \alpha_{c}\mathcal{L}_{center} .
\label{eq:shta_loss}
\end{equation}

\(\alpha_{as}\), \(\alpha_a\), \(\alpha_h\), and \(\alpha_c\) are loss
weights. \(\mathcal{L}_{assign}\) establishes token-level semantic assignments,
\(\mathcal{L}_{anchor}\) calibrates the learnable proxies with labeled-token references,
\(\mathcal{L}_{hard}\) corrects selected hard-token assignments, and
\(\mathcal{L}_{center}\) stabilizes class-level semantic centers. This SHTA objective is plugged
into the overall training objective in Eq.~\eqref{eq:total_objective} as an auxiliary
representation loss and supervises the base encoder-decoder through gradients during training.
At inference, the SHTA branch is removed, and the original segmentation head directly maps
\(F_A\) to \(\hat{Y}\).

\section{Experiments}

\subsection{Experimental Settings}

We evaluate SHTA on Synapse and AMOS CT under 3D volumetric semi-supervised segmentation protocols~\cite{synapse2015btcv,ji2022amos}. Synapse contains 30 abdominal CT scans with 13 foreground organs; following the common split, 20 scans are used for training, 4 for validation, and 6 for testing, with 20\% of the training scans treated as labeled data. AMOS is used as a secondary benchmark under the 5\% labeled setting. 

All internal comparisons keep the original framework architecture, base objective, and inference pathway unchanged; training is augmented only by the auxiliary SHTA loss in Eq.~\eqref{eq:total_objective}. SHTA is attached only during training, its hyperparameters are selected by validation performance, and the branch is removed for testing. All variants use case-wise inference on the held-out test set and follow the same volumetric evaluation protocol within each benchmark.

\subsection{Overall Performance}

We first evaluate whether SHTA improves segmentation performance under paired same-protocol comparisons across representative SSL frameworks.

\begin{table*}[!t]
\centering
\tiny
\caption{Comparison on Synapse under the 20\% labeled setting. Reported SSL rows provide external references from consistency-based, reliable-supervision, and hard-region/distribution-aware methods; same-protocol rows compare SHTA on CPS, URPC, GA-CPS, and MagicNet, covering prediction consistency, uncertainty-aware learning, imbalance-aware supervision, and structure-aware optimization.}
\label{tab:main_comparison}
\renewcommand{\arraystretch}{1.05}
\setlength{\tabcolsep}{1.6pt}
\resizebox{0.98\textwidth}{!}{
\begin{tabular}{l|l|ll|ccccccccccccc}
\toprule
\multirow{3}{*}{Group} & \multirow{3}{*}{Method} & \multicolumn{15}{c}{Synapse (20\% labeled)} \\[-1.5pt]
\cmidrule(lr){3-17}
&  & \multicolumn{2}{c|}{\textbf{Overall}} & \multicolumn{4}{c|}{\cellcolor{headbg}Large} & \multicolumn{5}{c|}{\cellcolor{medbg}Medium} & \multicolumn{4}{c}{\cellcolor{tailbg}Small} \\[-1.5pt]
&  & Dice $\uparrow$ & ASD $\downarrow$ & \cellcolor{headbg}Sp & \cellcolor{headbg}RK & \cellcolor{headbg}LK & \cellcolor{headbg}Li & \cellcolor{medbg}St & \cellcolor{medbg}Ao & \cellcolor{medbg}IVC & \cellcolor{medbg}PVSV & \cellcolor{medbg}Pa & \cellcolor{tailbg}GB & \cellcolor{tailbg}Eso & \cellcolor{tailbg}RAG & \cellcolor{tailbg}LAG \\[-1.5pt]
\midrule
\multirow{9}{*}{Reported SSL} & \makecell[l]{UA-MT \pubyear{2019}~\cite{ref_ssl_1}} & 42.16 & 15.48 & 59.8 & 64.9 & 64.0 & 77.7 & 37.8 & 61.0 & 46.0 & 33.3 & 26.9 & 35.3 & 34.1 & 12.3 & 18.1 \\[-1.5pt]
& \makecell[l]{URPC \pubyear{2022}~\cite{ref_ssl_2}} & 44.93 & 27.44 & 67.0 & 64.2 & 67.2 & 83.1 & 45.5 & 67.4 & 54.4 & 46.7 & 0.0 & 36.1 & 0.0 & 29.4 & 35.2 \\[-1.5pt]
& \makecell[l]{CPS \pubyear{2021}~\cite{ref_ssl_3}} & 41.08 & 20.37 & 56.1 & 60.3 & 59.4 & 73.8 & 32.4 & 65.7 & 52.1 & 31.1 & 25.5 & 33.3 & 25.4 & 6.2 & 18.4 \\[-1.5pt]
\cmidrule(lr){2-17}
& \makecell[l]{BCP \pubyear{2023}~\cite{bai2023bcp}} & 43.57 & 28.12 & 62.3 & 66.1 & 62.5 & 79.4 & 38.2 & 64.3 & 49.1 & 35.6 & 22.4 & 34.8 & 18.7 & 15.2 & 22.8 \\[-1.5pt]
& \makecell[l]{ABD \pubyear{2024}~\cite{chi2024abd}} & 49.10 & 22.45 & 68.7 & 70.4 & 68.2 & 84.2 & 42.1 & 69.8 & 54.7 & 41.2 & 28.6 & 39.5 & 25.3 & 21.8 & 29.4 \\[-1.5pt]
\cmidrule(lr){2-17}
& \makecell[l]{GA \pubyear{2024}~\cite{ref_imbalance_4}} & 66.45 & 4.58 & 78.9 & 85.5 & 87.2 & 86.9 & 56.2 & 83.4 & 70.3 & 57.4 & \textbf{49.1} & 50.0 & 49.1 & 38.3 & 71.6 \\[-1.5pt]
& \makecell[l]{SKCDF \pubyear{2025}~\cite{yin2025skcdf}} & 58.21 & 5.97 & 77.1 & 77.9 & 71.2 & 88.6 & 51.6 & 80.9 & 58.9 & 48.8 & 33.0 & 34.1 & 50.4 & 38.3 & 45.9 \\[-1.5pt]
& \makecell[l]{DHC \pubyear{2023}~\cite{wang2023dhc}} & 49.53 & 13.89 & 68.1 & 69.6 & 71.1 & 76.8 & 43.8 & 70.8 & 57.4 & 43.2 & 27.0 & 42.3 & 44.9 & 23.4 & 22.7 \\[-1.5pt]
& \makecell[l]{DCA \pubyear{2025}~\cite{wang2025dca}} & \textbf{69.90} & \textbf{2.66} & \textbf{87.3} & \textbf{87.8} & \textbf{90.6} & \textbf{91.3} & \textbf{66.4} & \textbf{87.0} & \textbf{76.7} & \textbf{64.0} & 48.5 & \textbf{59.6} & \textbf{63.6} & \textbf{46.1} & \textbf{72.7} \\
\midrule
\multirow{8}{*}{\makecell[l]{Same-protocol\\frameworks}} & \makecell[l]{URPC \pubyear{2022}~\cite{ref_ssl_2}} & 35.27 & 36.48 & \cellcolor{headbg}68.52 & \cellcolor{headbg}56.78 & \cellcolor{headbg}65.01 & \cellcolor{headbg}73.74 & \cellcolor{medbg}33.18 & \cellcolor{medbg}54.35 & \cellcolor{medbg}54.95 & \cellcolor{medbg}17.94 & \cellcolor{medbg}17.38 & \cellcolor{tailbg}16.67 & \cellcolor{tailbg}0.00 & \cellcolor{tailbg}0.00 & \cellcolor{tailbg}0.00 \\
& \makecell[l]{URPC \pubyear{2022}~\cite{ref_ssl_2} + SHTA} & 42.33\,\metricchange{+7.06} & 23.84\,\metricchange{-12.64} & \cellcolor{headbg}72.02 & \cellcolor{headbg}81.03 & \cellcolor{headbg}68.26 & \cellcolor{headbg}79.26 & \cellcolor{medbg}39.17 & \cellcolor{medbg}63.80 & \cellcolor{medbg}51.45 & \cellcolor{medbg}18.12 & \cellcolor{medbg}17.50 & \cellcolor{tailbg}0.26 & \cellcolor{tailbg}0.00 & \cellcolor{tailbg}38.47 & \cellcolor{tailbg}20.96 \\
\cmidrule(lr){2-17}
& \makecell[l]{CPS \pubyear{2021}~\cite{ref_ssl_3}} & 66.29 & 5.44 & \cellcolor{headbg}85.47 & \cellcolor{headbg}90.12 & \cellcolor{headbg}88.36 & \cellcolor{headbg}92.74 & \cellcolor{medbg}64.80 & \cellcolor{medbg}79.71 & \cellcolor{medbg}79.13 & \cellcolor{medbg}\textbf{66.80} & \cellcolor{medbg}45.49 & \cellcolor{tailbg}26.73 & \cellcolor{tailbg}40.24 & \cellcolor{tailbg}44.74 & \cellcolor{tailbg}57.38 \\
& \makecell[l]{CPS \pubyear{2021}~\cite{ref_ssl_3} + SHTA} & 67.50\,\metricchange{+1.21} & 3.32\,\metricchange{-2.12} & \cellcolor{headbg}88.18 & \cellcolor{headbg}91.66 & \cellcolor{headbg}89.26 & \cellcolor{headbg}\textbf{93.69} & \cellcolor{medbg}64.38 & \cellcolor{medbg}78.05 & \cellcolor{medbg}81.76 & \cellcolor{medbg}65.43 & \cellcolor{medbg}\textbf{49.42} & \cellcolor{tailbg}25.44 & \cellcolor{tailbg}38.66 & \cellcolor{tailbg}49.22 & \cellcolor{tailbg}\textbf{62.35} \\
\cmidrule(lr){2-17}
& \makecell[l]{MagicNet \pubyear{2023}~\cite{chen2023magicnet}} & 65.86 & \textbf{3.08} & \cellcolor{headbg}81.95 & \cellcolor{headbg}90.95 & \cellcolor{headbg}\textbf{90.19} & \cellcolor{headbg}90.75 & \cellcolor{medbg}57.22 & \cellcolor{medbg}80.49 & \cellcolor{medbg}80.25 & \cellcolor{medbg}63.93 & \cellcolor{medbg}42.34 & \cellcolor{tailbg}26.13 & \cellcolor{tailbg}42.38 & \cellcolor{tailbg}48.27 & \cellcolor{tailbg}61.31 \\
& \makecell[l]{MagicNet \pubyear{2023}~\cite{chen2023magicnet} + SHTA} & 65.98\,\metricchange{+0.12} & 3.68\,\metricchange{+0.60} & \cellcolor{headbg}84.87 & \cellcolor{headbg}\textbf{91.89} & \cellcolor{headbg}88.44 & \cellcolor{headbg}90.47 & \cellcolor{medbg}56.63 & \cellcolor{medbg}\textbf{80.63} & \cellcolor{medbg}80.11 & \cellcolor{medbg}63.78 & \cellcolor{medbg}39.96 & \cellcolor{tailbg}\textbf{30.35} & \cellcolor{tailbg}44.67 & \cellcolor{tailbg}47.05 & \cellcolor{tailbg}58.95 \\
\cmidrule(lr){2-17}
& \makecell[l]{GA-CPS \pubyear{2024}~\cite{ref_imbalance_4}} & 66.26 & 5.69 & \cellcolor{headbg}84.19 & \cellcolor{headbg}91.26 & \cellcolor{headbg}89.13 & \cellcolor{headbg}92.79 & \cellcolor{medbg}\textbf{65.23} & \cellcolor{medbg}79.11 & \cellcolor{medbg}80.81 & \cellcolor{medbg}65.23 & \cellcolor{medbg}47.43 & \cellcolor{tailbg}22.78 & \cellcolor{tailbg}40.45 & \cellcolor{tailbg}46.00 & \cellcolor{tailbg}57.01 \\
& \makecell[l]{GA-CPS \pubyear{2024}~\cite{ref_imbalance_4} + SHTA} & \textbf{68.56\,\metricchange{+2.30}} & 3.78\,\metricchange{-1.91} & \cellcolor{headbg}\textbf{88.43} & \cellcolor{headbg}91.23 & \cellcolor{headbg}88.84 & \cellcolor{headbg}93.05 & \cellcolor{medbg}64.22 & \cellcolor{medbg}78.50 & \cellcolor{medbg}\textbf{82.18} & \cellcolor{medbg}66.01 & \cellcolor{medbg}47.13 & \cellcolor{tailbg}24.56 & \cellcolor{tailbg}\textbf{49.37} & \cellcolor{tailbg}\textbf{56.13} & \cellcolor{tailbg}61.62 \\
\bottomrule
\end{tabular}}
\end{table*}

As shown in Table~\ref{tab:main_comparison}, the Synapse results provide the primary paired evidence for SHTA under matched implementation and evaluation settings. On already competitive baselines, SHTA gives clear paired improvements. For GA-CPS, mean Dice increases from 66.26 to 68.56 (\(+2.30\)) and ASD decreases from 5.69 to 3.78 (\(-1.91\)), with notable gains on weak organs such as esophagus from 40.45 to 49.37 (\(+8.92\)) and RAG from 46.00 to 56.13 (\(+10.13\)). CPS also improves mean Dice from 66.29 to 67.50 (\(+1.21\)) and ASD from 5.44 to 3.32 (\(-2.12\)), indicating that SHTA can refine a strong prediction-consistency baseline rather than only rescuing weak models. For MagicNet, which already includes a structure-aware design, the mean-Dice change is modest, from 65.86 to 65.98 (\(+0.12\)), while ASD changes from 3.08 to 3.68 (\(+0.60\)), suggesting that the marginal room for improvement is smaller. On the weaker URPC baseline, SHTA mainly acts as a recovery mechanism, improving mean Dice from 35.27 to 42.33 (\(+7.06\)) and restoring weak classes such as RAG from 0.00 to 38.47 (\(+38.47\)) and LAG from 0.00 to 20.96 (\(+20.96\)). Overall, Table~\ref{tab:main_comparison} shows consistent paired benefits across different SSL paradigms, with the largest practical gains appearing on weak and ambiguity-prone organs.
\begin{table*}[!t]
\centering
\tiny
\caption{Comparison on AMOS under the 5\% labeled setting. Reported SSL rows follow the grouping in Table~\ref{tab:main_comparison}; same-protocol rows compare each baseline with its SHTA variant under matched settings.}
\label{tab:amos_comparison}
\renewcommand{\arraystretch}{0.92}
\setlength{\tabcolsep}{1.3pt}
\renewcommand\cellgape{\Gape[0pt]}
\resizebox{0.98\textwidth}{!}{
\begin{tabular}{l|l|ll|ccccccccccccccc}
\toprule
\multirow{3}{*}{Group} & \multirow{3}{*}{Method} & \multicolumn{17}{c}{AMOS (5\% labeled)} \\[-1.5pt]
\cmidrule(lr){3-19}
&  & \multicolumn{2}{c|}{\textbf{Overall}} & \multicolumn{5}{c|}{\cellcolor{headbg}Large} & \multicolumn{6}{c|}{\cellcolor{medbg}Medium} & \multicolumn{4}{c}{\cellcolor{tailbg}Small} \\[-1.5pt]
&  & Dice $\uparrow$ & ASD $\downarrow$ & \cellcolor{headbg}Sp & \cellcolor{headbg}RK & \cellcolor{headbg}LK & \cellcolor{headbg}Li & \cellcolor{headbg}Bla & \cellcolor{medbg}St & \cellcolor{medbg}Ao & \cellcolor{medbg}PVC & \cellcolor{medbg}Pa & \cellcolor{medbg}Duo & \cellcolor{medbg}Pro/Ute & \cellcolor{tailbg}GB & \cellcolor{tailbg}Eso & \cellcolor{tailbg}RAG & \cellcolor{tailbg}LAG \\[-1.5pt]
\midrule
\multirow{9}{*}{Reported SSL} & \makecell[l]{UA-MT \pubyear{2019}~\cite{ref_ssl_1}} & 20.26 & 71.67 & 48.2 & 31.7 & 22.2 & 81.2 & 29.7 & 0.0 & 23.3 & 0.0 & 0.0 & 18.1 & 31.6 & 0.0 & 0.0 & 0.0 & 0.0 \\[-1.5pt]
& \makecell[l]{URPC \pubyear{2022}~\cite{ref_ssl_2}} & 25.68 & 72.74 & 66.7 & 38.2 & 56.8 & 85.3 & 44.5 & 0.0 & 33.1 & 0.0 & 5.1 & 35.2 & 33.2 & 0.0 & 0.0 & 0.0 & 0.0 \\[-1.5pt]
& \makecell[l]{CPS \pubyear{2021}~\cite{ref_ssl_3}} & 33.55 & 41.21 & 62.8 & 55.2 & 45.4 & 91.1 & 40.7 & 35.9 & 41.9 & 8.8 & 14.5 & 18.4 & 35.8 & 0.0 & 0.0 & 0.0 & 0.0 \\[-1.5pt]
\cmidrule(lr){2-19}
& \makecell[l]{BCP \pubyear{2023}~\cite{bai2023bcp}} & 50.23 & 18.45 & 72.4 & 58.1 & 52.7 & 92.3 & 35.1 & 38.2 & 48.9 & 18.7 & 22.1 & 22.8 & 32.9 & 8.4 & 12.5 & 8.4 & 5.2 \\[-1.5pt]
& \makecell[l]{ABD \pubyear{2024}~\cite{chi2024abd}} & 55.67 & 15.32 & 76.8 & 64.3 & 58.9 & \textbf{93.7} & 42.3 & 42.5 & 55.4 & 24.3 & 28.7 & 29.4 & 38.7 & 14.6 & 18.9 & 14.6 & 11.8 \\[-1.5pt]
\cmidrule(lr){2-19}
& \makecell[l]{GA \pubyear{2024}~\cite{ref_imbalance_4}} & 68.43 & 3.11 & 81.4 & 92.4 & 90.8 & 89.1 & 71.6 & 33.5 & 79.1 & 66.7 & 48.7 & 38.3 & 47.9 & 50.3 & 53.3 & 50.3 & \textbf{61.4} \\[-1.5pt]
& \makecell[l]{SKCDF \pubyear{2025}~\cite{yin2025skcdf}} & 64.27 & \textbf{1.45} & 79.5 & 72.1 & 67.6 & 93.3 & 45.9 & 59.8 & \textbf{85.4} & 41.8 & 50.9 & 32.2 & 26.4 & 46.4 & 60.7 & 46.4 & 37.8 \\[-1.5pt]
& \makecell[l]{DCA \pubyear{2025}~\cite{wang2025dca}} & \textbf{73.20} & 1.78 & \textbf{82.9} & \textbf{92.8} & \textbf{91.4} & 92.5 & \textbf{72.7} & 64.0 & 83.9 & \textbf{68.1} & \textbf{52.2} & \textbf{46.1} & \textbf{51.1} & \textbf{51.8} & \textbf{61.4} & \textbf{63.7} & 48.5 \\[-1.5pt]
& \makecell[l]{DHC \pubyear{2023}~\cite{wang2023dhc}} & 48.61 & 10.71 & 62.8 & 69.5 & 59.2 & 85.2 & 41.4 & \textbf{66.0} & 67.9 & 37.0 & 30.9 & 29.1 & 36.7 & 31.4 & 13.2 & 31.4 & 10.6 \\
\midrule
\multirow{8}{*}{\makecell[l]{Same-protocol\\frameworks}} & \makecell[l]{URPC \pubyear{2022}~\cite{ref_ssl_2}} & 41.72 & 36.07 & \cellcolor{headbg}68.51 & \cellcolor{headbg}62.74 & \cellcolor{headbg}68.27 & \cellcolor{headbg}41.26 & \cellcolor{headbg}0.00 & \cellcolor{medbg}83.73 & \cellcolor{medbg}42.42 & \cellcolor{medbg}68.51 & \cellcolor{medbg}54.74 & \cellcolor{medbg}32.81 & \cellcolor{medbg}0.00 & \cellcolor{tailbg}0.00 & \cellcolor{tailbg}24.58 & \cellcolor{tailbg}59.03 & \cellcolor{tailbg}19.17 \\
& \makecell[l]{URPC \pubyear{2022}~\cite{ref_ssl_2} + SHTA} & 44.32\,\metricchange{+2.60} & 14.99\,\metricchange{-21.08} & \cellcolor{headbg}64.58 & \cellcolor{headbg}53.45 & \cellcolor{headbg}57.88 & \cellcolor{headbg}21.93 & \cellcolor{headbg}44.06 & \cellcolor{medbg}82.21 & \cellcolor{medbg}39.52 & \cellcolor{medbg}66.80 & \cellcolor{medbg}53.24 & \cellcolor{medbg}32.56 & \cellcolor{medbg}23.91 & \cellcolor{tailbg}21.13 & \cellcolor{tailbg}24.05 & \cellcolor{tailbg}47.43 & \cellcolor{tailbg}32.05 \\
\cmidrule(lr){2-19}
& \makecell[l]{CPS \pubyear{2021}~\cite{ref_ssl_3}} & 61.93 & 10.83 & \cellcolor{headbg}80.56 & \cellcolor{headbg}83.24 & \cellcolor{headbg}81.87 & \cellcolor{headbg}43.86 & \cellcolor{headbg}51.65 & \cellcolor{medbg}\textbf{87.50} & \cellcolor{medbg}60.78 & \cellcolor{medbg}\textbf{80.33} & \cellcolor{medbg}\textbf{67.40} & \cellcolor{medbg}59.55 & \cellcolor{medbg}39.77 & \cellcolor{tailbg}39.55 & \cellcolor{tailbg}39.82 & \cellcolor{tailbg}68.74 & \cellcolor{tailbg}44.28 \\
& \makecell[l]{CPS \pubyear{2021}~\cite{ref_ssl_3} + SHTA} & 62.20\,\metricchange{+0.27} & 11.57\,\metricchange{+0.74} & \cellcolor{headbg}80.18 & \cellcolor{headbg}80.15 & \cellcolor{headbg}80.34 & \cellcolor{headbg}49.76 & \cellcolor{headbg}49.66 & \cellcolor{medbg}87.21 & \cellcolor{medbg}62.89 & \cellcolor{medbg}79.70 & \cellcolor{medbg}64.15 & \cellcolor{medbg}\textbf{59.96} & \cellcolor{medbg}38.35 & \cellcolor{tailbg}38.22 & \cellcolor{tailbg}42.09 & \cellcolor{tailbg}\textbf{71.45} & \cellcolor{tailbg}\textbf{48.87} \\
\cmidrule(lr){2-19}
& \makecell[l]{MagicNet \pubyear{2023}~\cite{chen2023magicnet}} & 63.47 & 5.86 & \cellcolor{headbg}80.52 & \cellcolor{headbg}83.45 & \cellcolor{headbg}83.95 & \cellcolor{headbg}87.61 & \cellcolor{headbg}66.61 & \cellcolor{medbg}60.38 & \cellcolor{medbg}\textbf{84.32} & \cellcolor{medbg}72.62 & \cellcolor{medbg}57.04 & \cellcolor{medbg}42.04 & \cellcolor{medbg}47.86 & \cellcolor{tailbg}48.35 & \cellcolor{tailbg}50.35 & \cellcolor{tailbg}48.40 & \cellcolor{tailbg}38.53 \\
& \makecell[l]{MagicNet \pubyear{2023}~\cite{chen2023magicnet} + SHTA} & \textbf{63.58\,\metricchange{+0.11}} & \textbf{5.11\,\metricchange{-0.75}} & \cellcolor{headbg}\textbf{80.60} & \cellcolor{headbg}\textbf{84.00} & \cellcolor{headbg}\textbf{87.20} & \cellcolor{headbg}87.40 & \cellcolor{headbg}\textbf{69.20} & \cellcolor{medbg}59.60 & \cellcolor{medbg}83.80 & \cellcolor{medbg}72.30 & \cellcolor{medbg}56.80 & \cellcolor{medbg}42.10 & \cellcolor{medbg}\textbf{48.00} & \cellcolor{tailbg}\textbf{52.20} & \cellcolor{tailbg}\textbf{52.80} & \cellcolor{tailbg}50.80 & \cellcolor{tailbg}39.70 \\
\cmidrule(lr){2-19}
& \makecell[l]{GA-CPS \pubyear{2024}~\cite{ref_imbalance_4}} & 57.88 & 8.85 & \cellcolor{headbg}77.69 & \cellcolor{headbg}78.31 & \cellcolor{headbg}77.38 & \cellcolor{headbg}\textbf{88.48} & \cellcolor{headbg}67.69 & \cellcolor{medbg}50.91 & \cellcolor{medbg}76.18 & \cellcolor{medbg}61.13 & \cellcolor{medbg}49.50 & \cellcolor{medbg}32.61 & \cellcolor{medbg}45.56 & \cellcolor{tailbg}46.87 & \cellcolor{tailbg}39.28 & \cellcolor{tailbg}41.83 & \cellcolor{tailbg}34.75 \\
& \makecell[l]{GA-CPS \pubyear{2024}~\cite{ref_imbalance_4} + SHTA} & 59.83\,\metricchange{+1.95} & 8.60\,\metricchange{-0.25} & \cellcolor{headbg}80.08 & \cellcolor{headbg}79.54 & \cellcolor{headbg}77.09 & \cellcolor{headbg}87.79 & \cellcolor{headbg}69.08 & \cellcolor{medbg}52.14 & \cellcolor{medbg}77.56 & \cellcolor{medbg}65.09 & \cellcolor{medbg}54.20 & \cellcolor{medbg}35.75 & \cellcolor{medbg}44.22 & \cellcolor{tailbg}49.86 & \cellcolor{tailbg}48.73 & \cellcolor{tailbg}42.21 & \cellcolor{tailbg}34.17 \\
\bottomrule
\end{tabular}}
\end{table*}

Table~\ref{tab:amos_comparison} further tests whether the same behavior holds under the more challenging AMOS 5\% labeled setting. SHTA again improves mean Dice for all same-protocol frameworks: URPC from 41.72 to 44.32 (\(+2.60\)), CPS from 61.93 to 62.20 (\(+0.27\)), MagicNet from 63.47 to 63.58 (\(+0.11\)), and GA-CPS from 57.88 to 59.83 (\(+1.95\)). The strongest effect appears on URPC, where ASD decreases from 36.07 to 14.99 (\(-21.08\)) and missing weak structures are partially recovered, such as GB from 0.00 to 21.13 (\(+21.13\)) and LAG from 19.17 to 32.05 (\(+12.88\)). For the stronger AMOS baselines, the overall gains are smaller but still concentrated on difficult structures. MagicNet improves GB from 48.35 to 52.20 (\(+3.85\)), esophagus from 50.35 to 52.80 (\(+2.45\)), and RAG from 48.40 to 50.80 (\(+2.40\)); GA-CPS improves PVC from 61.13 to 65.09 (\(+3.96\)), pancreas from 49.50 to 54.20 (\(+4.70\)), and esophagus from 39.28 to 48.73 (\(+9.45\)). For CPS, SHTA still improves mean Dice from 61.93 to 62.20 (\(+0.27\)), while the ASD remains in a comparable range. Taken together, Tables~\ref{tab:main_comparison} and~\ref{tab:amos_comparison} show that SHTA provides consistent paired accuracy improvements across datasets and frameworks, with boundary-related changes varying by framework and dataset.

\subsection{Computational Overhead}

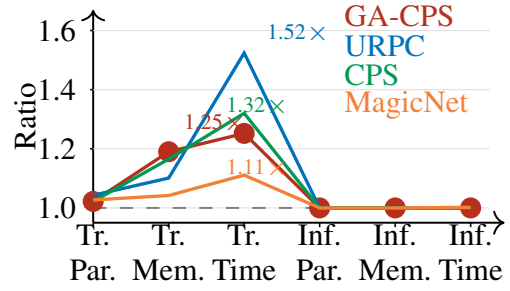
\begin{figure}[!t]
\centering
\resizebox{0.76\columnwidth}{!}{%
\begin{tikzpicture}[x=0.60cm,y=2.35cm]
\scriptsize
\draw[->,line width=0.45pt] (0,0) -- (5.45,0);
\draw[->,line width=0.45pt] (0,0) -- (0,0.72);
\foreach \x/\lab in {0/{Tr.\\Par.},1/{Tr.\\Mem.},2/{Tr.\\Time},3/{Inf.\\Par.},4/{Inf.\\Mem.},5/{Inf.\\Time}} {
  \draw[line width=0.35pt] (\x,0) -- (\x,-0.015);
  \node[align=center,font=\scriptsize] at (\x,-0.10) {\lab};
}
\foreach \y/\lab in {0.05/1.0,0.25/1.2,0.45/1.4,0.65/1.6} {
  \draw[line width=0.35pt] (-0.04,\y) -- (0,\y);
  \node[anchor=east,font=\scriptsize] at (-0.08,\y) {\lab};
  \draw[gray!25,line width=0.25pt] (0,\y) -- (5.25,\y);
}
\draw[gray,dashed,line width=0.45pt] (0,0.05) -- (5.25,0.05);
\draw[BrickRed,line width=0.75pt,mark=*] plot coordinates
  {(0,0.072) (1,0.240) (2,0.302) (3,0.050) (4,0.050) (5,0.050)};
\draw[RoyalBlue,line width=0.75pt,mark=square*] plot coordinates
  {(0,0.093) (1,0.151) (2,0.574) (3,0.050) (4,0.050) (5,0.050)};
\draw[ForestGreen,line width=0.75pt,mark=triangle*] plot coordinates
  {(0,0.072) (1,0.215) (2,0.370) (3,0.050) (4,0.050) (5,0.050)};
\draw[Orange,line width=0.75pt,mark=diamond*] plot coordinates
  {(0,0.077) (1,0.092) (2,0.161) (3,0.050) (4,0.050) (5,0.052)};
\node[font=\tiny, text=BrickRed] at (1.62,0.335) {1.25$\times$};
\node[font=\tiny, text=RoyalBlue] at (2.72,0.64) {1.52$\times$};
\node[font=\tiny, text=ForestGreen] at (2.18,0.395) {1.32$\times$};
\node[font=\tiny, text=Orange] at (2.18,0.185) {1.11$\times$};
\node[anchor=west,font=\scriptsize,text=BrickRed] at (3.18,0.70) {GA-CPS};
\node[anchor=west,font=\scriptsize,text=RoyalBlue] at (3.18,0.60) {URPC};
\node[anchor=west,font=\scriptsize,text=ForestGreen] at (3.18,0.50) {CPS};
\node[anchor=west,font=\scriptsize,text=Orange] at (3.18,0.40) {MagicNet};
\node[rotate=90,font=\scriptsize] at (-0.9,0.36) {Ratio};
\end{tikzpicture}%
}
\caption{Training and inference overhead ratios across different frameworks, normalized by their respective baselines. Annotated values denote training-time ratios.}
\label{fig:overhead_ratio}
\end{figure}
As summarized in Fig.~\ref{fig:overhead_ratio}, the overhead of SHTA is mainly limited to training. Training parameter ratios remain close to the baseline (about 1.02--1.04$\times$), memory increases moderately (1.04--1.19$\times$), and the annotated training-time ratios range from 1.11$\times$ on MagicNet to 1.52$\times$ on URPC, with GA-CPS at 1.25$\times$ and CPS at 1.32$\times$. By contrast, inference parameters and memory stay at 1.00$\times$, and inference time remains about 1.00$\times$ across frameworks. This confirms that the auxiliary semantic branch affects optimization cost but not the deployed inference pathway. We next examine whether the three semantic operations are individually responsible for the observed gains.

\begin{figure*}[!t]
\centering
\includegraphics[width=0.86\textwidth]{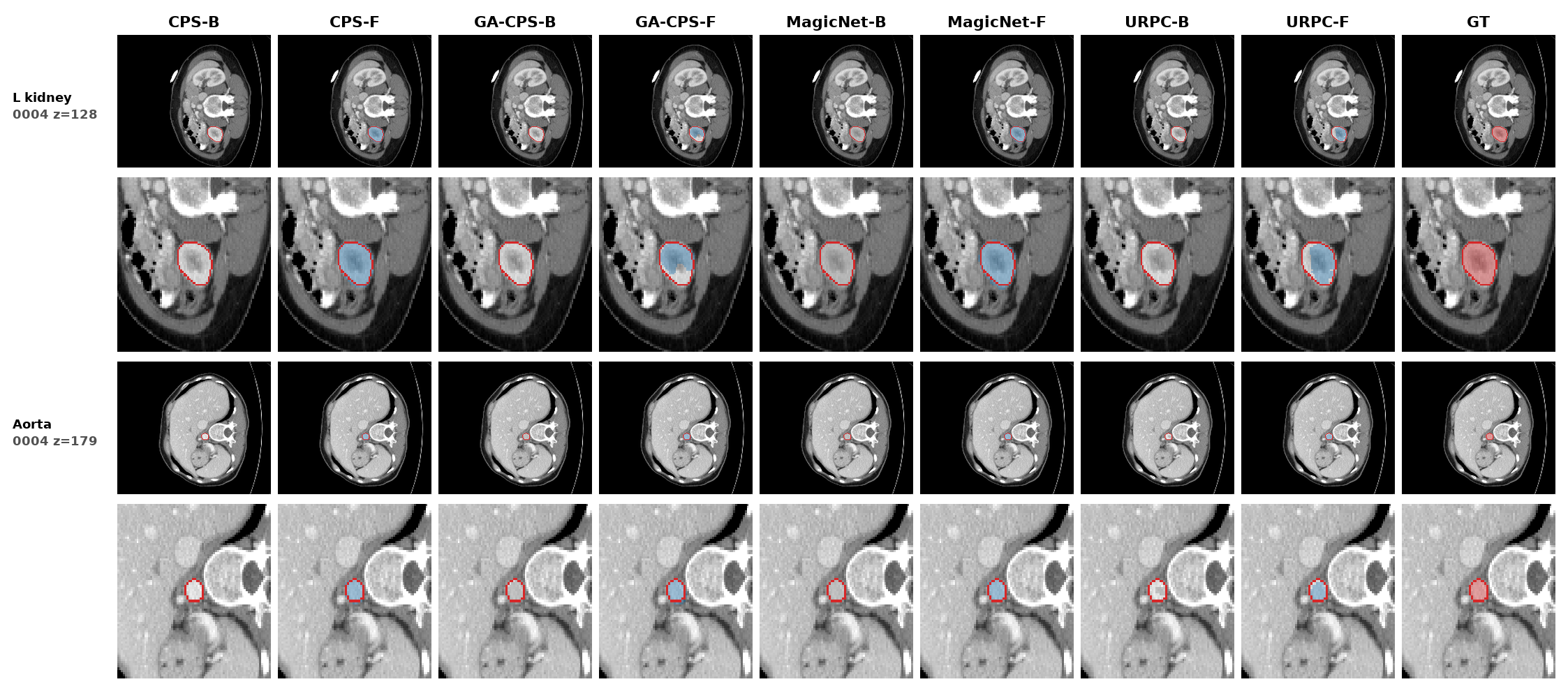}
\caption{Qualitative comparison across SSL frameworks on Synapse. B and F denote baseline and full SHTA variants; red contours denote GT boundaries and blue regions denote predictions. The displayed cases include left kidney and aorta examples across CPS, GA-CPS, MagicNet, and URPC.}
\label{fig:cross_framework_generality}
\end{figure*}

\subsection{Ablation Study}

After the overall comparison, we isolate the contribution of each SHTA component. The ablation uses the same framework, optimizer, data split, and evaluation protocol, while enabling one semantic module at a time. This isolates each method component without changing the prediction model, training schedule, or test-time inference pathway.

\begin{table*}[!t]
\centering
\tiny
\caption{Component ablation on 3D Synapse and AMOS. Assign, Hard, and Center denote Semantic Assignment, Hard Token Refinement, and Semantic Center Alignment, respectively.}
\label{tab:component_overview}
\renewcommand{\arraystretch}{0.9}
\setlength{\tabcolsep}{3.2pt}
\resizebox{0.68\textwidth}{!}{
\begin{tabular}{ccc|ccccc|ccccc}
\toprule
\multirow{2}{*}{\makecell{Assign}} & \multirow{2}{*}{\makecell{Hard}} & \multirow{2}{*}{\makecell{Center}} & \multicolumn{5}{c|}{3D Synapse} & \multicolumn{5}{c}{3D AMOS} \\
\cmidrule(lr){4-8}\cmidrule(lr){9-13}
&  &  & \makecell{Large\\$\uparrow$} & \makecell{Medium\\$\uparrow$} & \makecell{Small\\$\uparrow$} & \textbf{\makecell[l]{Mean\\Dice $\uparrow$}} & \textbf{\makecell[l]{ASD\\$\downarrow$}} & \makecell{Large\\$\uparrow$} & \makecell{Medium\\$\uparrow$} & \makecell{Small\\$\uparrow$} & \textbf{\makecell[l]{Mean\\Dice $\uparrow$}} & \textbf{\makecell[l]{ASD\\$\downarrow$}} \\
\midrule
-- & -- & -- & 89.34 & 67.56 & 41.56 & 66.26 & 5.69 & 77.91 & 52.65 & 40.68 & 57.88 & 8.85 \\
\rowcolor{lightblue2}
$\checkmark$ & -- & -- & 90.44 & 66.54 & 40.52 & 65.88 & \textbf{3.28} & 77.58 & 53.30 & 42.50 & 58.51 & \textbf{7.82} \\
\rowcolor{lightblue2}
-- & $\checkmark$ & -- & 90.40 & 66.73 & 44.73 & 67.24 & 4.03 & 78.74 & 53.94 & 42.50 & 59.15 & 8.34 \\
\rowcolor{lightblue2}
-- & -- & $\checkmark$ & \textbf{90.98} & \textbf{67.83} & 45.28 & 68.02 & 6.46 & \textbf{79.19} & 53.36 & 42.73 & 59.13 & 8.61 \\
\rowcolor{lightpink1}
$\checkmark$ & $\checkmark$ & $\checkmark$ & 90.39 & 67.61 & \textbf{47.92} & \textbf{68.56} & 3.78 & 78.72 & \textbf{54.83} & \textbf{43.74} & \textbf{59.83} & 8.60 \\
\bottomrule
\end{tabular}}
\end{table*}

Table~\ref{tab:component_overview} shows that the three stages play different roles. Semantic Assignment mainly provides the semantic scaffold for later correction by calibrating token-to-class assignment with labeled guidance, reducing ASD from 5.69 to 3.28 on Synapse (\(-2.41\)) and from 8.85 to 7.82 on AMOS (\(-1.03\)). Hard Token Refinement then operates on selected difficult tokens rather than all features, giving clearer Dice gains: mean Dice increases from 66.26 to 67.24 on Synapse (\(+0.98\)) and from 57.88 to 59.15 on AMOS (\(+1.27\)); Synapse small-organ Dice also rises from 41.56 to 44.73 (\(+3.17\)). Semantic Center Alignment further stabilizes class-level representation geometry, increasing mean Dice from 66.26 to 68.02 on Synapse (\(+1.76\)) and from 57.88 to 59.13 on AMOS (\(+1.25\)). With all three stages integrated, SHTA achieves the largest mean-Dice gains, from 66.26 to 68.56 on Synapse (\(+2.30\)) and from 57.88 to 59.83 on AMOS (\(+1.95\)), together with the strongest small-organ gains from 41.56 to 47.92 on Synapse (\(+6.36\)) and from 40.68 to 43.74 on AMOS (\(+3.06\)), while keeping ASD below the baseline. These quantitative trends motivate a closer look at whether the corrected predictions also correspond to more consistent local and token-level semantics.

\begin{figure*}[!t]
\centering
\includegraphics[width=0.70\textwidth]{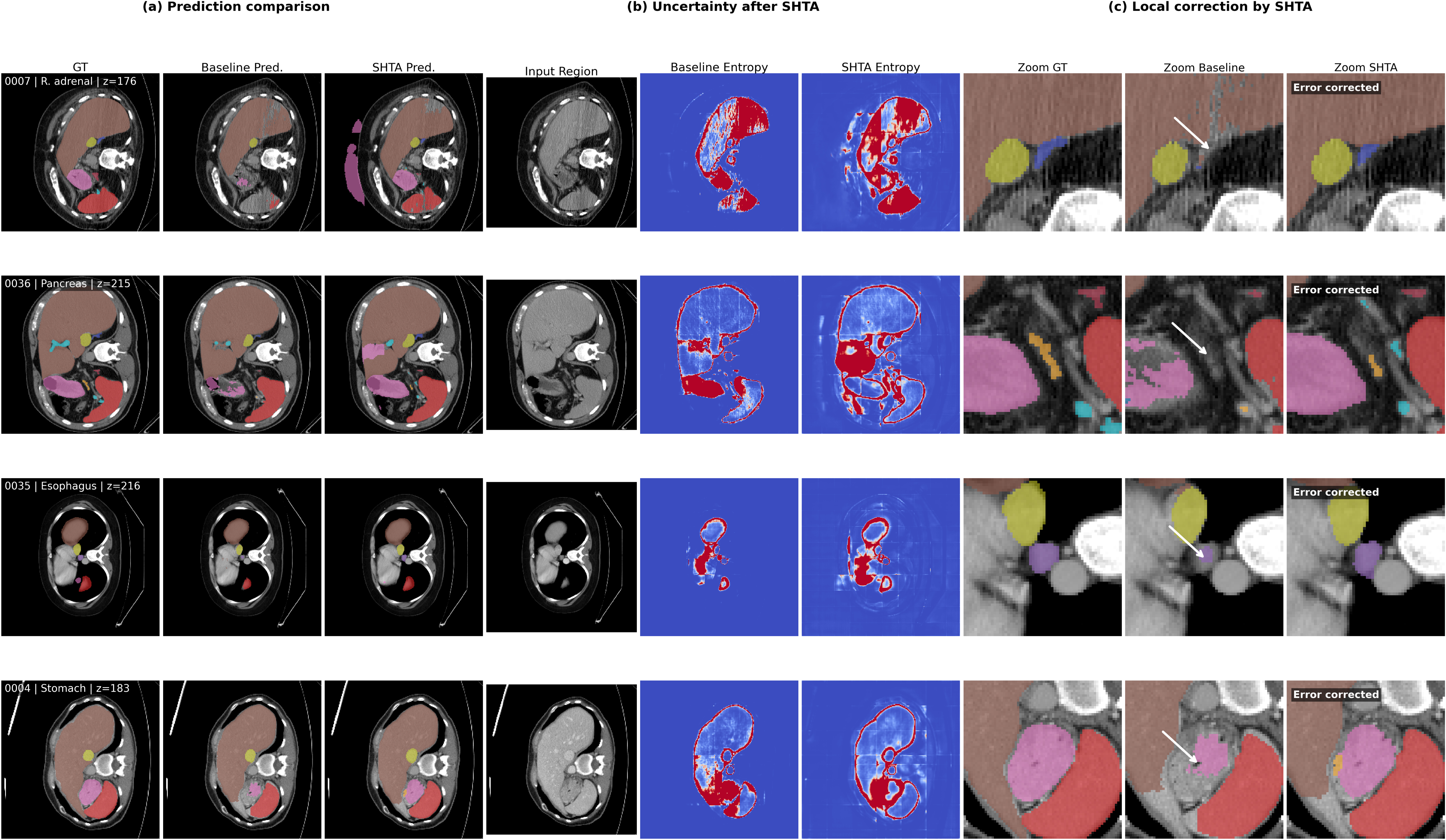}
\caption{Region-level ambiguity reduction on Synapse. Zoomed regions compare GT, baseline predictions, SHTA predictions, and entropy maps; arrows mark the local regions used for visual comparison.}
\label{fig:semantic_ambiguity_reduction}
\end{figure*}

\begin{figure}[!t]
\centering
\includegraphics[width=\columnwidth]{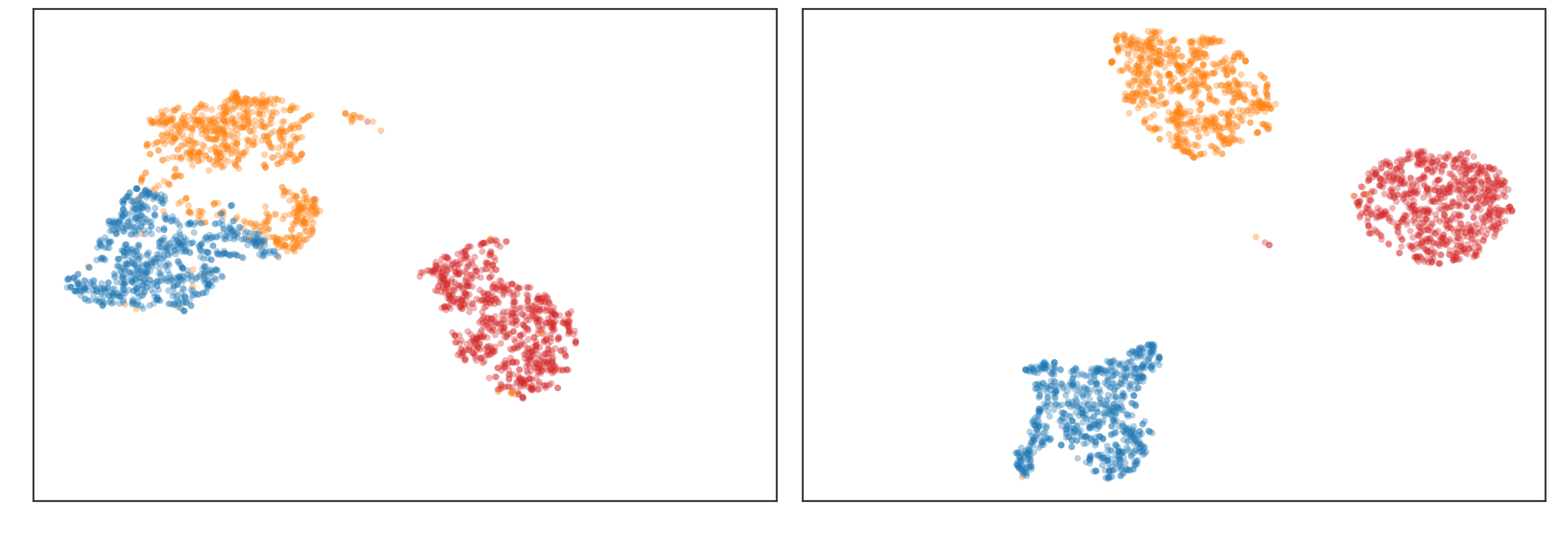}
\caption{Representation-level semantic consistency on Synapse. Decoder embeddings before the segmentation classifier are shown for left kidney, pancreas, and right adrenal gland.}
\label{fig:semantic_tsne}
\end{figure}

\begin{figure}[!t]
\centering
\includegraphics[width=0.8\columnwidth]{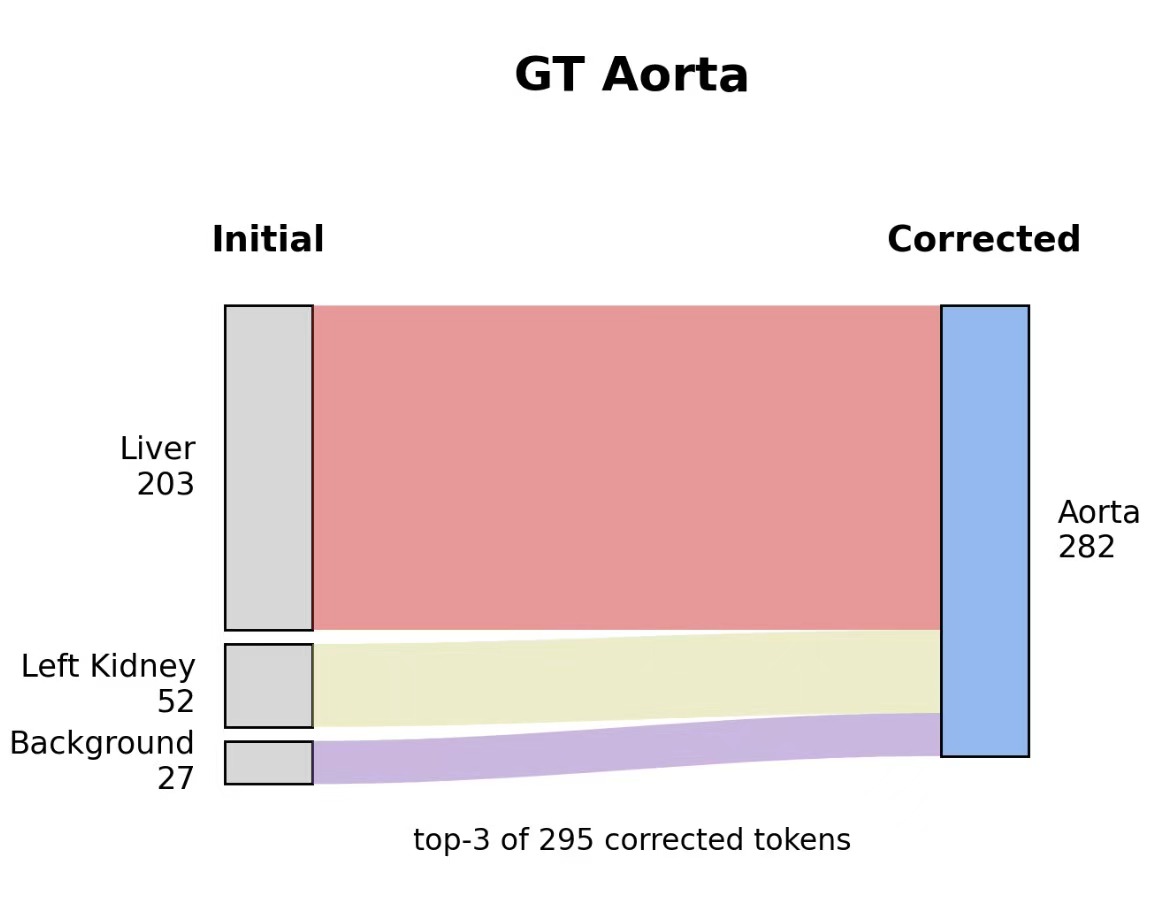}
\caption{Hard-token assignment refinement on Synapse. Flows connect initial proxy assignments and GT-derived dominant classes; flow width denotes token count.}
\label{fig:semantic_assignment_flow}
\end{figure}

\subsection{Semantic Consistency Analysis}

Beyond the aggregate Dice and ASD scores, we further examine the error patterns and token behavior behind the improvements. Fig.~\ref{fig:cross_framework_generality}
  shows two representative organs, left kidney and aorta, across CPS, GA-CPS, MagicNet, and URPC. In these cases, several baselines almost miss the target organ or produce
  only very weak activation inside the red GT contour. After adding SHTA, the target regions become visibly activated and closer to the GT location, especially for the
  small aorta region. This suggests that SHTA helps recover weak organs that are not reliably recognized by the original SSL frameworks.

  Fig.~\ref{fig:semantic_ambiguity_reduction} further shows where these corrections occur. The right adrenal gland, pancreas, esophagus, and stomach examples focus on local
  regions where the baseline prediction is missing, fragmented, or mixed with nearby structures. In the zoomed views, SHTA restores more coherent target-organ responses and
  suppresses part of the local false activation. The entropy maps further indicate where the model remains uncertain: high-entropy responses in the baseline appear around
  the same local regions where the target organs are missed or confused, while SHTA makes these uncertain regions lower and more localized, matching the corrected areas in
  the zoomed predictions.

  The representation and assignment analyses give a finer explanation. In Fig.~\ref{fig:semantic_tsne}, the three displayed classes correspond to left kidney, pancreas, and
  right adrenal gland. In the baseline feature space, pancreas and adrenal-gland tokens are partially mixed with neighboring class distributions, and the left-kidney tokens
  are less compact. After SHTA, the three organ-specific token groups become more clearly separated and internally tighter before the final segmentation classifier.
  Fig.~\ref{fig:semantic_assignment_flow} shows the token-level correction behind this change. Tokens with GT-dominant stomach, left kidney, aorta, and spleen semantics are
  initially attracted to competing classes such as liver, kidney, stomach, or background, but are corrected to their target classes after refinement. For example, aorta
  tokens are reassigned from liver (203), left kidney (52), and background (27) into 295 corrected aorta tokens.

  Table~\ref{tab:mechanism} further rules out the explanation that SHTA simply selects more hard tokens. Hard Token Refinement only and Full SHTA use the same 3.76\% valid
  hard-token ratio, about 2.47k tokens per iteration. Under this matched supply, adding semantic assignment and center alignment reduces the correction loss from 0.129 to
  0.054. Thus, SHTA improves how selected hard tokens are semantically corrected, rather than increasing the number of selected tokens, linking the visual recovery of weak
  organs to the intended representation-level correction.

\begin{table}[!t]
\centering
\tiny
\caption{Matched-hard mechanism readouts on Synapse.}
\label{tab:mechanism}
\renewcommand{\arraystretch}{1}
\setlength{\tabcolsep}{2.0pt}
\resizebox{\columnwidth}{!}{
\begin{tabular}{lcccc}
\toprule
\makecell{Setting} & \makecell{Semantic\\scaffold} & \makecell{Hard-token\\supply} & \makecell{Correction\\loss $\downarrow$} & \makecell{Center\\readout $\downarrow$} \\
\midrule
\makecell[l]{Hard Token\\Refinement only} & inactive & \makecell{3.76\%\\2.47k} & 0.129 & inactive \\
\rowcolor{lightpink1}
Full SHTA & \makecell{assignment loss 0.325\\anchor loss 0.039} & \makecell{3.76\%\\2.47k} & \textbf{0.054} & \makecell{center loss 0.034\\center gap 0.238} \\
\bottomrule
\end{tabular}}
\end{table}

\section{Conclusion}

We presented SHTA, a training-time semantic representation branch for semi-supervised medical image segmentation. Instead of redesigning the segmentation framework or adding inference-time prediction modules, SHTA refines intermediate representations through semantic assignment, hard-token correction, and semantic center alignment. This design treats weak-class failure as a semantic consistency problem after difficult evidence selection, while preserving the original prediction objective and deployment pathway of the base SSL framework.

Experiments on Synapse and AMOS show that SHTA improves paired same-protocol baselines across representative SSL paradigms, with the most evident gains on weak and ambiguity-prone anatomical structures. The ablation and semantic analyses further indicate that the improvement comes from complementary effects: assignment provides class-guided semantic scaffolding, hard-token refinement corrects ambiguous token assignments, and center alignment stabilizes class-level representation geometry. At the same time, the observed gains remain framework-, dataset-, and metric-dependent, especially for ASD, so the evidence supports improved semantic consistency and weak-organ behavior rather than uniform dominance across all organs or metrics. Since the auxiliary branch is removed after training, SHTA introduces only training-time overhead and keeps the deployed inference architecture unchanged.

\begingroup
\footnotesize
\setlength{\itemsep}{0pt}
\setlength{\parskip}{0pt}

\endgroup

\end{document}